\documentclass[conference]{IEEEtran}
\IEEEoverridecommandlockouts
\usepackage{cite}
\usepackage{amsmath,amssymb,amsfonts}
\usepackage{algorithmic}
\usepackage{graphicx}
\usepackage{textcomp}
\usepackage{multirow}

\usepackage{subfigure}
\usepackage[table,xcdraw]{xcolor}

\usepackage{tikz}

\AtBeginDocument{%
 \providecommand\BibTeX{{%
 \normalfont B\kern-0.5em{\scshape i\kern-0.25em b}\kern-0.8em\TeX}}}
 
\newcommand\copyrighttext{%
  \footnotesize \textcopyright \the\year{} IEEE. Personal use of this material is permitted.  Permission from IEEE must be obtained for all other uses, in any current or future media, including reprinting/republishing this material for advertising or promotional purposes, creating new collective works, for resale or redistribution to servers or lists, or reuse of any copyrighted component of this work in other works.}

\newcommand\copyrightnotice{%
\begin{tikzpicture}[remember picture,overlay]
\node[anchor=south,yshift=10pt] at (current page.south) {\fbox{\parbox{\dimexpr0.75\textwidth-\fboxsep-\fboxrule\relax}{\copyrighttext}}};
\end{tikzpicture}%
}

\newcolumntype{M}{>{$\displaystyle}c<{$}}
\newcommand\AddLabel[1]{\refstepcounter{equation}(\theequation)\label{#1}}
\newcolumntype{L}{>{\collectcell\AddLabel}r<{\endcollectcell}}

\begin{document}
\title{Benchmarking Advanced Text Anonymisation Methods: A Comparative Study on Novel and Traditional Approaches
}

\author{Dimitris Asimopoulos\IEEEauthorrefmark{1}, Ilias Siniosoglou\IEEEauthorrefmark{1}\IEEEauthorrefmark{2}, Vasileios Argyriou\IEEEauthorrefmark{3}, Thomai Karamitsou\IEEEauthorrefmark{4},  Eleftherios Fountoukidis\IEEEauthorrefmark{4}, \\Sotirios K. Goudos\IEEEauthorrefmark{5}, Ioannis D. Moscholios\IEEEauthorrefmark{6}, Konstantinos E. Psannis\IEEEauthorrefmark{7} and Panagiotis Sarigiannidis\IEEEauthorrefmark{1}\IEEEauthorrefmark{2}

\thanks{\IEEEauthorrefmark{1} I. Siniosoglou, D. Asimopoulos and P. Sarigiannidis are with the R\&D Department, MetaMind Innovations P.C., Kozani, Greece - \texttt{E-Mail: \{isiniosoglou, dasimopoulos, psarigiannidis\}@metamind.gr}}

\thanks{\IEEEauthorrefmark{2} I. Siniosoglou and P. Sarigiannidis are with the Department of Electrical and Computer Engineering, University of Western Macedonia, Kozani, Greece - \texttt{E-Mail: \{isiniosoglou, psarigiannidis\}@uowm.gr}}

\thanks{\IEEEauthorrefmark{3} V. Argyriou is with the Department of Networks and Digital Media, Kingston University, Kingston upon Thames, United Kingdom - \texttt{E-Mail: vasileios.argyriou@kingston.ac.uk}}

\thanks{\IEEEauthorrefmark{4}  T. Karamitsou and E. Fountoukidis are with Sidroco Holdings Ltd., Nicosia, Cyprus - \texttt{E-Mail: \{tkaramitsou, efountoukidis\}@sidroco.com}}

\thanks{\IEEEauthorrefmark{5} S. K. Goudos is with the Physics Department, Aristotle University of Thessaloniki, Thessaloniki, Greece   - \texttt{E-Mail: sgoudo@physics.auth.gr}}

\thanks{\IEEEauthorrefmark{6} I. D. Moscholios is with the Department of Informatics and Telecommunications Department, University of Peloponnese, Tripoli, Greece   - \texttt{E-Mail: idm@uop.gr}}

\thanks{\IEEEauthorrefmark{7} K. E. Psannis is with the Department of Applied Informatics, School of Information Sciences, University of Macedonia, Thessaloniki, Greece   - \texttt{E-Mail: kpsannis@uom.edu.gr}}
}

\maketitle
\copyrightnotice

\begin{abstract}
In the realm of data privacy, the ability to effectively anonymise text is paramount. With the proliferation of deep learning and, in particular, transformer architectures, there is a burgeoning interest in leveraging these advanced models for text anonymisation tasks. This paper presents a comprehensive benchmarking study comparing the performance of transformer-based models and Large Language Models(LLM) against traditional architectures for text anonymisation. Utilising the CoNLL-2003 dataset, known for its robustness and diversity, we evaluate several models. Our results showcase the strengths and weaknesses of each approach, offering a clear perspective on the efficacy of modern versus traditional methods. Notably, while modern  models exhibit advanced capabilities in capturing contextual nuances, certain traditional architectures still keep high performance. This work aims to guide researchers in selecting the most suitable model for their anonymisation needs, while also shedding light on potential paths for future advancements in the field.
\end{abstract}


\section{Introduction}
\label{Introduction}
Ensuring the privacy and security of data in today's interconnected world has emerged as a critical challenge. Textual data, a significant fraction of the digital ecosystem, frequently contains sensitive information. As such, the ability to effectively anonymise text is a key component of modern data protection paradigms. This paper presents a detailed benchmarking of various text anonymisation methodologies, focusing on the comparison between the modern models,such as transformers,LLM, and traditional architectures.

Efficient data anonymisation solutions are receiving increased focus, as they pose a critical issue for the protection of people’s and organisations’ privacy. This is all the more apparent because of the reality that data creation and collection are increasing exponentially in modern digital Cloud-Edge ecosystems. To aid in this process, data obfuscation has emerged as an integral part of the data handling, curration and processing pipeline in order to actively protect sensitive information, while it is seen that is very precise. It is a process that involves encrypting, erasing or otherwise scrambling sensitive information identifiers that link an individual or process with the data they belong to. Particularly the encryption process involves substitution or removal of crucial text and numerical data, in order to be unable to identify sensitive information without authorization due to lack of specific context. In is important to note though that, depending on the application of, this procedure can still make available significant information about the data, like distribution, statistics, and so on, but with the sensitive information redacted. 

With data volumes and depth in the information age increasing and becoming more complicated for organizations to handle, privacy protection requirements are becoming all the more stricter due to legal sanctions for failure to protect private user information. The emergence of these challenges calls for the development of cutting-edge anonymisation tools that are capable of overcoming the different data terrains while following the privacy principles and rights of each individual. Primarily, first approaches to that kind of task used the rule-based and dictionary-based techniques. While there has been improvement in data privacy, the dynamic and fast changing data privacy issues remain a big puzzle that may require more advanced solutions.

Machine learning and NLP technologies, such as Conditional Random Fields (CRF) \cite{Zhiheng2015}, Long Short-Term Memory (LSTM) networks, and ELMo for Named Entity Recognition (NER), served as pioneering approaches, illuminating potential pathways for data anonymisation. Alas, the rapid advancements in the field brought the era of transformer models. Transformers pose a significant innovation in deep learning offering advanced capabilities, like:

\begin{itemize}
    \item {\textbf{Parallel Processing:} Transformers are characterized by the ability to process data simultaneously which improves the efficiency and management of larger datasets.}
    \item {\textbf{Attention Mechanisms:} Attention at the heart of Transformer is one of the main reasons for the model capability to dynamically shift its focus from one part of the data to another, making the model capable of learning patterns and relationships between complex information.}
     \item{\textbf{Scalability:} Due to being designed to be trainable on a variety of model sizes, transformers are inherently scalable.}

\end{itemize}

Due to these characteristics, LLMs can effectively be used in the field of Named Entity Recognition.  The state-of-the-art LLMS are equipped with modules that can detect language subtleties hence making them effective in the realm of anonymization tasks. Not only that, the structure of their neural networks is very sophisticated as it entails continuous learning capacity while accommodating to new patterns of languages that occasionally change. Since this enables the LLMs to recognize real entites with more precision in an environment, which is changing dynamically, their use is worthwhile.  The integration of LLMs into NER tasks shows their ability
 to detect sensitive information in a level where they can be compared with the traditional models and transformers.


This paper aims to gap the responsibility of classical anonymization strategies, transformer-based models and LLMs, while evaluating their efficacy for real-world applications. This work undertakes to evaluate the performance of state-of-the-art AI models, like GPT2\cite{Qu2020}, BERT\cite{Vakili2022}, and ELECTR\cite{Catelli2021} by finetuning these models on the CoNLL-2003 dataset \cite{Tjong2003}, which is known for its heterogeneity and robustness. Thereon this work dives into the ML models such as LSTMs\cite{Siniosoglou2021_a}, CRF, and other more traditional models to provide a holistic comparative analysis, providing insights for their results and efficasy on the task of data anonymisation. 

\section{Literature Review}
\label{Literature Review}

The work of \cite{Kotei2023} focuses on the developments that have been made in NLP as a result of transformer networks applied through self-supervised learning, highlighting the value of transfer learning in reducing model overfitting by introducing a huge unlabeled dataset. This work veers into the role of pre-trained models such as BERT\cite{Vakili2022} and GPT that have brought about a paradigm shift in NLP by introducing an approach which does not require extensive labeled datasets, thus improving the efficiency of downstream activities. Pilan et al. \cite{Pilan2022}, present a special corpus and assessment framework designed to evaluate different text de-anonymization algorithms. It differentiates between direct indicators, such as personal names and social security numbers, and quasi-indicators, such as demographics, that when in combination, could result in individual identification. The paper is aimed at demonstrating the trade-off between the confidentiality risk level and the data utility level in the process of anonymization. In this paper, Nikoletos et al.\cite{Nikoletos2023} highlight the growing demand for data security while the number of online users increases, bringing forth the issues of safeguarding critical information from misuse. They focuses on the issues of data protection, which comprises of legal, ethical and technical aspects and which urges the use of automated tools in collecting and anonymizing sensitive data. The work suggests a new process of fully automatic NLP-based system that will enable both high degree of efficiency and effectiveness, and will be suitable for various data sets across different domains. Furthermore, Pierre Lison et al. \cite{Pilan2023}, explore automated text anonymization, essential for securely sharing sensitive information. The presented work reviews current methods from natural language processing and privacy-preserving data publishing, highlighting their benefits, limitations, and lack of interaction. Key challenges identified include handling semantic inferences, balancing disclosure risk against data utility, and evaluating anonymization quality. The paper advocates for advancements beyond traditional sequence labeling models to include explicit disclosure risk measures, aiming to improve the anonymization process's effectiveness. On the other hand, in \cite{Patsakis2023} the authors search the efficacy of text anonymization methods in the context of modern AI capabilities, particularly focusing on the challenge of balancing privacy protection with data utility. It questions the adequacy of current anonymization techniques to mitigate re-identification risks amidst the advancements in AI and big data analytics. Through an experiment with GPT on anonymized texts of notable individuals, the study evaluates the potential for re-identification by AI, leading to a proposal for a novel approach that leverages Large Language Models to enhance text anonymity.

\subsection{Overview of Text Anonymisation}
\label{Overview of Text Anonymisation}

Data anonymization has become an irreplaceable method in the sphere of cyber-security as it helps obfuscate confidential documents safeguarding sensitive information. This is enhanced by the fact that both sensitive and private data remain under the threat of cyber attacks, or being used for illegal purposes.  In this ever-evolving cyber-security landscape, practitioners and researchers have crafted an amalgam of strategies to proficiently pinpoint sensitive data and subsequently anonymise them. Among these, the application of NER principles takes center stage \cite{nasar2021named}, \cite{baigang2023review}. The strength of NER lies in its capability to make objective assessments of entities, differentiating between personal and organisational references. By doing so, it plays a pivotal role in highlighting data that may be considered sensitive or private. Subsequently, in data anonymization, after the critical identification stage there comes the neutralisation of the anonymized data. The strategy for neutralization is intricately designed which is goal-oriented and accounting for the structure of data and the specifications of cyber security projects. The most widely adopted techniques are:
\begin{itemize}
    \item {\textbf{Removal:} A straightforward approach, this method eliminates references to confidential data, substituting them with generic placeholders. The outcome is data cleansed of its sensitive elements.}
    \item {\textbf{Categorization:} More nuanced than removal, this technique uses labels instead of direct references. It offers a general insight into the nature of the anonymised data without divulging specifics.}
     \item{\textbf{Pseudonymisation:} This method replaces sensitive records with alternatives that, while different, belong to the same category of data. It's especially relevant for contexts where the type of data needs to be retained, but specific details must be obscured.}

\end{itemize}

\begin{table}[htb]
  \centering
  \caption{Example of anonymisation process}
  \label{Example of anonymisation process}
  \resizebox{\linewidth}{!}{%
  \begin{tabular}{|p{2cm}|l|l|}
    \hline
    \textbf{Methods} & \textbf{Original Data} & \textbf{Transformed Data} \\
    \hline
    Removal & John Smith works at HSBC Bank & \textless REF\textgreater{} works at \textless REF\textgreater \\
    \hline
    Categorisation & John Smith works at HSBC Bank & \textless PERSON\textgreater{} works at \textless LOCATION\textgreater \\
    \hline
    Pseudonymisation & John Smith works at HSBC Bank & Peter Green works at NatWest Bank \\
    \hline
  \end{tabular}
}
\end{table}

The challenges in data anonymisation are manifold. The inherent subjectivity associated with what constitutes 'sensitive' information, combined with a scarcity of extensively annotated training datasets across sectors, has propelled the rise of Natural Language Processing (NLP) techniques \cite{raj2021anonymization}, \cite{li2020survey}. NLP, with its robust framework for handling NER tasks, is further augmented by the adaptability of machine learning. This synergy ensures that solutions are not only effective but are also customised to the nuances of each specific case, thereby bolstering the reliability and efficacy of the entire anonymisation process.
To further push the boundaries and help in future anonymisation directions, our current research is centered on providing an extended comparative analysis between novel models widely used for anonymisation tasks. This integration aims to enhance the precision and depth of NER tasks within the sphere of anonymisation, setting the stage for even more refined outcomes.

\section{Methodology}
\label{Methodology}

\begin{figure}[!httb]
\begin{center}
 \includegraphics[width=0.95\linewidth]{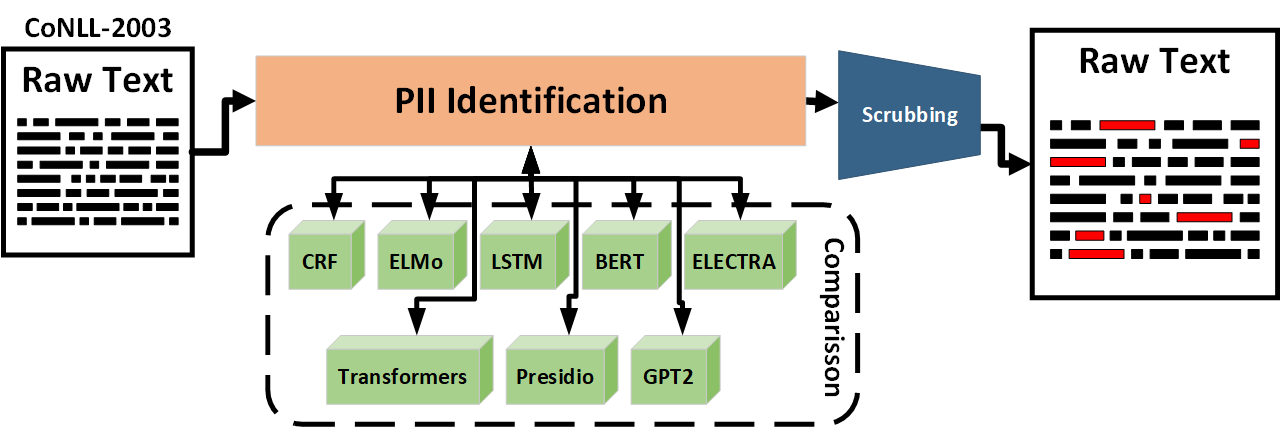}
\end{center}
\caption{Anonymisation Pipeline}
\label{img:anon_process}
\end{figure}

\subsection{Traditional Models}
\label{Traditional Models}
On of the main pillars of this work is to evaluate the NER identification process using traditional models currently used, namely, CRF, LSTM and ELMo. The CRF method, a statistical modeling technique made for structured prediction, is adopted, as it is considered one of the state-of-the-art for sequence labeling tasks. CRFs models see data anonymization as a sequence labeling problem and hence precisely label and classify personal data entities (names, address, or dates) in the text sequences by putting them under an anonymized descriptor. On the other hand, the success of Long Short Term Memory (LSTM), which is based on RNNs, is better understood as a result of its ability to make connections between long-term dependencies in sequential data, being able to temporaly identify connection of PIIs in text. To eliminate private identifiers from source materials, LSTM  networks are trained to distinguish patterns and contexts that disclose the presence of sensitive information, thus enabling the effective redacting thereof, while preserving the meaning of the text. The last baseline model used for anonymisation is ELMo \cite{Taill2020}. ELMo, known for generating deep, contextualised word representations, is utilised to understand the nuanced meanings of words in different contexts. This understanding enables ELMo to discern between instances when PIIs should be anonymised and when similar words are used in non-sensitive contexts. Overall, each of these traditional models offers unique strengths in the anonymisation process providing a holistic approach to the anonymisation task guaranteeing effective protection of privacy and retaining the authenticity of the textual data.

\subsection{Transformers Models}
\label{Transformers Models}

In the methodology of applying transformer models to the task of data anonymisation, we primarily focus on three advanced architectures: BERT\cite{Vakili2022}, ELECTRA\cite{Catelli2021}, and a custom Transformer model. BERT, a bi-directional approach that has been recognized for its deep structure, is leveraged to understand each word in a sentence, and it consequently makes it possible to detect and modify such data. This is designated by adjusting BERT, which is annotated with personal data, by training it to acknowledge and replace them with neutral placeholders. ELECTRA, it distinguished between real and replaced tokens in text, increasing in this way its performance in detecting any different or context-related inforamtion. This is crucial for comprehension of various approaches to shield different data categories with confidentiality. Finally, the Transforemr model which is based on the transformer architecture, has been finetuned for anonymisation. In this regard, it combines the core strengths of transformers using of the a hybrid approach with bidirectional context understanding together with efficient replacement strategies. The model gets trained and customized on a wide range of texts featuring different data types and formats so that the anonymisation process is consistently effective and extensive. Aggregated, these models represent an outstanding approach to data anonymization, inherently providing predictive insights into the challenges of preserving textual data privacy.

\subsection{Microsoft Presidio Model}
\label{Microsoft Presidio Model}

\begin{figure}[!httb]
\begin{center}
 \includegraphics[width=0.7\linewidth]{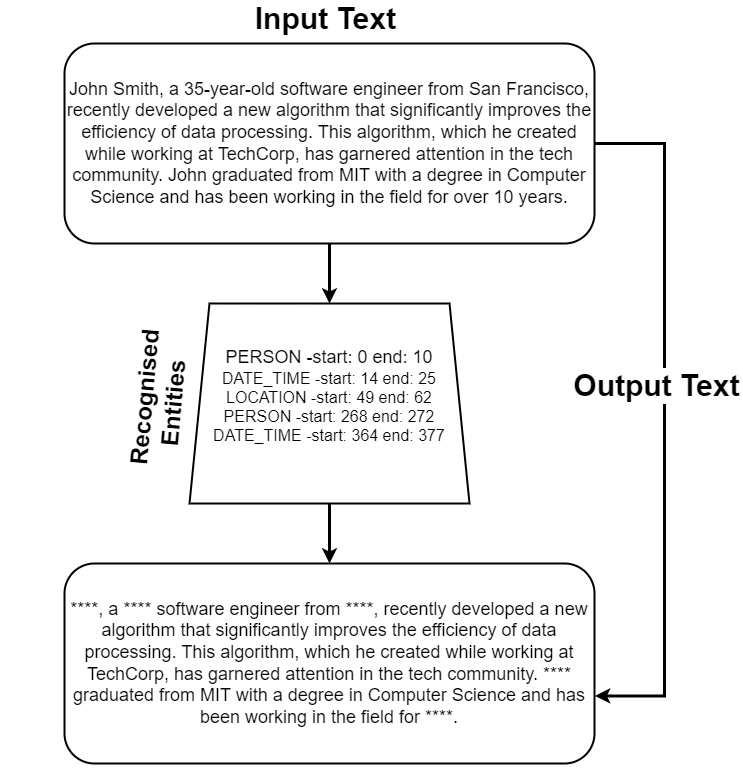}
\end{center}
\caption{Presidio Anonymisation}
\label{img:Presidio Methodology}
\end{figure}

In the domain of data anonymisation, Microsoft Presidio emerges as a robust, purpose-built tool that leverages advanced machine learning techniques to detect and anonymise sensitive information in text. Presidio operates by first identifying a wide range of personal data types, such as names, addresses, social security numbers, and credit card information, using a combination of predefined and customisable detectors. These detectors are grounded in pattern recognition, checksum validation, and contextual analysis, ensuring a high degree of accuracy in identifying sensitive data. Once identified, Presidio employs a series of anonymisation strategies, including substitution, redaction, and generalization, to effectively obscure the identified information as shown in Figure \ref{img:Presidio Methodology}. 

Data substitution involves replacing existing data with fabricated information, while data elimination entails removing data from the system entirely. On the other hand, data masking replaces sensitive data with more generalized information. Presidio stands out due to its exceptional capability to expand detection rules and anonymization methods, making it adaptable to a wide range of specific data privacy requirements.

\subsection{LLM Model}
\label{LLM Model}

In the realm of data anonymization, Large Language Models (LLMs) such as the GPT model introduce a groundbreaking approach. These models, having undergone training on diverse datasets, possess a deep understanding of various language idioms, contexts, and semantic nuances. LLMs, particularly the GPT family, are extensively utilized in anonymization tasks due to their adeptness in comprehending and modifying text while preserving its original meaning and ensuring individuality. The methodology involves fine-tuning the GPT model using a dataset annotated with sensitive information, enabling the model to identify and replace specific types of sensitive data accurately. Unlike other substitution or masking methods, GPT models can generate contextually relevant replacements for sensitive details, maintaining the coherence and readability of the text. This capability is particularly valuable in scenarios where anonymization entails substituting plausible, non-sensitive alternatives for the removed information. Additionally, GPT models excel in adapting to various text styles and formats, rendering them indispensable tools for diverse anonymization tasks across different domains. GPT-2, with its transformer architecture leveraging attention mechanisms, demonstrates superior performance in tasks such as translation, question-answering, and text summarization. For our study, we utilize the GPT-2 model for text anonymization, leveraging its proficiency in recognizing and handling sensitive data effectively.

\subsection{Dataset \& Preprocessing}
\label{Dataset & Reprocessing}
For the evaluation of the methods outlined in this work, we leverage the CoNLL-2003 datasets \cite{Tjong2003}. This dataset host a repertoire of tasks that go from the detection of individual entities to the in-depth analysis of words accompanied by the recognition of word to word relations. A vital aspect that shaped my choice for the CoNLL dataset was the richness and diversity in the dataset.To leverage the complete potential of the CoNLL dataset for advanced language processing tasks, a meticulous preprocessing regimen is essential. The following key steps were undertaken, 1) Tokenization, 2) IDS conversion, 3) Padding \& Truncating, 4) Splitting Data, and 5) Converting Data into Tensors.

The CoNLL dataset sentence undergoes systematic preprocessing through a series of defined steps to facilitate experimentation. Tokenization is initially employed, breaking down the entire textual content into smaller units known as tokens. These tokens are then converted into IDs that correspond to the indices of the model's vocabulary embedding. To address variability, sequences are equally padded and truncated to maintain the efficiency of the neural network architecture. Subsequently, the data is divided into separate portions for training, validation, and testing, facilitating model training and evaluation. Finally, to ensure alignment with the deep learning framework and optimize model performance, the data is normalized and converted into torch tensors, enabling efficient matrix operations and compatibility with the framework.

\section{Experimental Results}
\label{Experimental Results}
This section provides an in-dept comparative study of the abovementioned AI models on the task of NER recognition and PII deidentification.
\begin{table}[httb]
\centering
\caption{Performance Models in PII Identification}
\label{tab:all_model_comp}
\begin{tabular}{|l|l|l|l|}
\hline
\textbf{Model} & \textbf{Precision} & \textbf{Recall} & \textbf{F1} \\
\hline
\hline
\textbf{CRF}\cite{Zhiheng2015} & \textbf{0.93} & \textbf{0.93} & \textbf{0.93}\\
\hline
ELMo\cite{Taill2020} & 0.72 & 0.81 & 0.76  \\
\hline
LSTM & 0.93 & 0.92 & 0.92 \\
\hline
BERT\cite{Vakili2022} & 0.8 & 0.81 & 0.8\\
\hline
ELECTRA\cite{Catelli2021} & 0.74 & 0.77 & 0.75  \\
\hline
\textbf{Transformer} & \textbf{0.94} & \textbf{0.95} & \textbf{0.95} \\
\hline
Presidio\cite{Kotevski2022} & 0.83 & 0.88 & 0.85\\
\hline
GPT2\cite{Qu2020} & 0.70 & 0.79 & 0.71\\
\hline
\end{tabular}
\end{table}

\subsection{Performance of Traditional Models}
\label{Performance of Traditional Models}

\begin{figure}[!httb]
\begin{center}
 \includegraphics[width=0.65\linewidth]{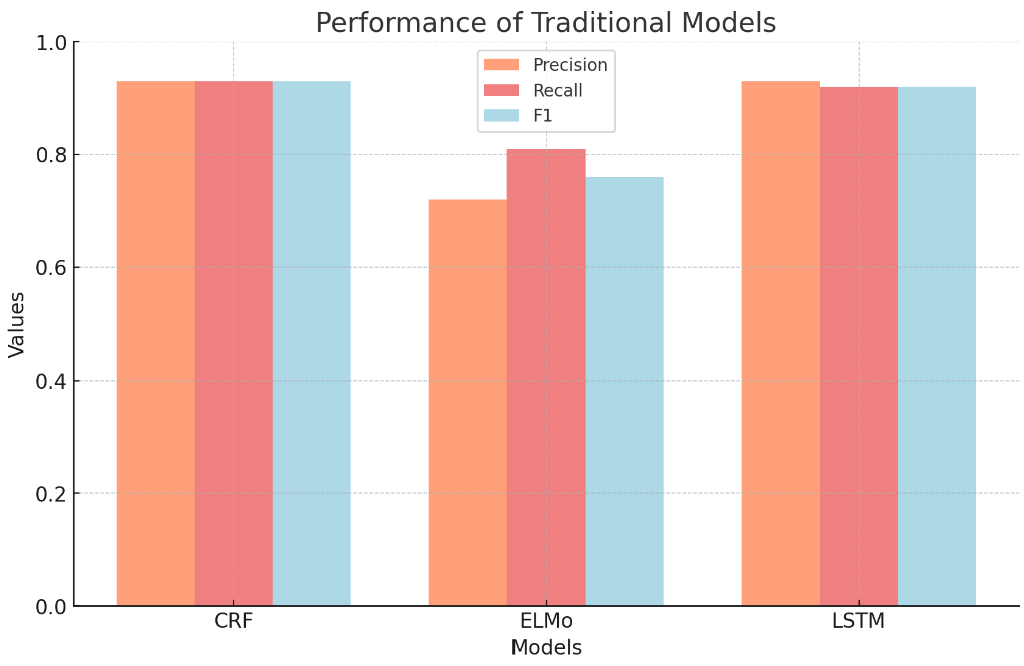}
\end{center}
\caption{Performance of Traditional Models}
\label{img:Performance of Traditional Models}
\end{figure}

Second in our study on anonymisation using traditional models, we compared the performance of CRF, ELMo, and LSTM as shown in Table \ref{tab:all_model_comp} and Figure \ref{img:Performance of Traditional Models}. The CRF model had the best performance, by achieving a precision, recall, and F1 score all at 0.93. Such uniformity across the different metrics reveals that the model has a balance and good performance in both of detecting and anonymise sensitive information. In contrast, ELMo had precision of 0.72, recall of 0.81, and an F1 score of 0.76.

A higher recall indicates ELMo's ability in identifying relevant cases, suggesting its effectiveness in this aspect. However, the precision reveals the presence of false positives in the predictions made by ELMo. On the other hand, the LSTM model exhibits performance comparable to CRF, with a precision of 0.93, a recall of 0.92, and an F1 score of 0.92. These results underscore the effectiveness of LSTM, showing as good performance  as CRF. Ultimately, among the three models evaluated, CRF and LSTM emerged as the top performers, although all models demonstrated competence in the anonymization task.

\subsection{Performance of Transformer Models}
\label{ Performance of Transformer Models}

\begin{figure}[!httb]
\begin{center}
 \includegraphics[width=0.65\linewidth]{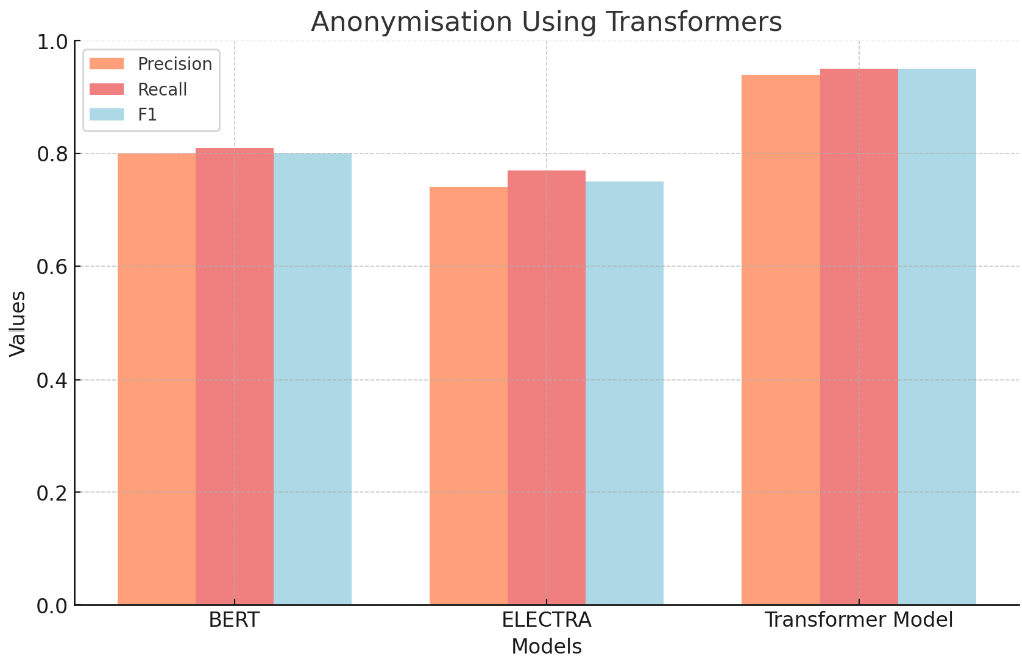}
\end{center}
\caption{Performance of Transformers Models}
\label{img:Performance of Transformers Models}
\end{figure}

Firstly, we measured the prowess of the variety of Transformers' models. As shown in Table \ref{tab:all_model_comp} and in figure \ref{img:Performance of Transformers Models}, the best results have been achieved by the transformer models. The ELECTRA module had a precision of 0.74 and a recall of 0.77. Consequently, its F1 score was 0.75. The outcomes of this experiment scrutinize that BERT has retained the ability to recognize and match the relevant terms, in contrast to the BERT model, which had better results as its precision was 0.8, and recall was 0.81 while the end result came to be an F1 score of 0.8. The comparison between BERT and ELECTRA verifies that the first learnt a little better than the second in terms of recalling the appropriate instances and classifying them properly.

Noteworthy is the Custom Model that achieved the highest accuracy of 0.94, recall of 0.95 and an F1 score of 0.95. This performance demonstrates the custom model's outstanding ability to extract the sensitive information from the dataset used, showing its ability on this anonymisation task. In summary, although all transformers models share the same architecture, the custom model outperformed the other two in the current evaluation.

\subsection{Performance of Microsoft Presidio}
\label{Performance of Microsoft Presidio}

Moreover, in our evaluation of anonymisation solutions, we assessed the capabilities of Microsoft's Presidio model.

Results presented in Table \ref{tab:all_model_comp} and Figure \ref{img:overall_performance} showed that Presidio is a good performer in this area. It attained a precision of 0.83, indicating that most of its predictions were correct or relevant. On the other hand, recall was at 0.88, meaning the model was effective at finding and capturing many cases which are relevant from the dataset used. The F1 score reveals a balanced performance exhibiting both recall and precision with an overall harmonization to achieve 0.85 respectively.This robust performance confirms Presidio's competency as an anonymization tool, showcasing its comprehensive and accurate coverage for data anonymization tasks.

\subsection{Performance of GPT2 Model}
\label{Performance of GPT2}

\begin{figure}[!httb]
\begin{center}
 \includegraphics[width=0.95\linewidth]{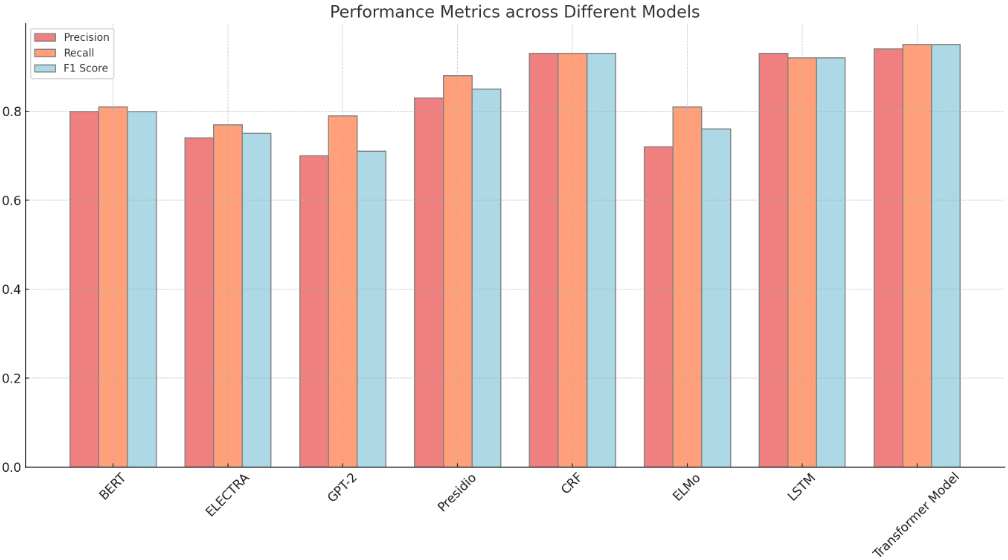}
\end{center}
\caption{Overall Performance of models}
\label{img:overall_performance}
\end{figure}

Table \ref{tab:all_model_comp} and Figure \ref{img:overall_performance} presents a succinct overview of the performance metrics for the GPT2 model in an anonymisation task. The GPT2 model exhibits a Precision of 0.70, indicating that 70\% of the model's identifications are correct. Its Recall is 0.79, suggesting that it successfully identifies 79\% of all relevant instances and the F1-score, which balances Precision and Recall, is 0.71, indicating a good balance between the precision and recall capabilities of the model. These metrics collectively suggest that the GPT2 model performs reasonably well in anonymising data, but there is room for improvement, especially in increasing precision without significantly sacrificing recall. 
\subsection{Comparative Analysis}
\label{Compartive Analysis}


In our comprehensive evaluation of various anonymisation models, we observed a diverse range of performances in Table \ref{tab:all_model_comp}. Starting with transformer models, BERT achieved a precision of 0.8, recall of 0.81, and an F1 score of 0.8. In comparison, ELECTRA recorded slightly lower values with a precision of 0.74, recall of 0.77, and an F1 score of 0.75. The custom Transformer Model surpassed both with  metrics of 0.94 across precision, recall, and F1 score, indicating an almost optimal balance between prediction accuracy and retrieval capability.
Shifting focus to Microsoft's Presidio, it showcased a robust performance, attaining a precision of 0.83, a recall of 0.88, and an F1 score of 0.85. These results underline Presidio's ability to blend accurate prediction with extensive instance retrieval. Finally, GPT2 model as a generative model achieved precision 0.70, recall 0.79 and F1 score 0.71 showing its ability to perform as good as the transformer models to NER and anonymisation tasks.

Among the traditional models, both CRF and LSTM demonstrated strikingly similar performances, with CRF scoring 0.93 across all metrics and LSTM closely trailing with 0.93 in precision, 0.92 in recall, and 0.92 for the F1 score. Their consistent scores across the board emphasise their reliability in the anonymisation task. In contrast, ELMo, although decent, lagged behind its peers with a precision of 0.72, recall of 0.81, and an F1 score of 0.76.

In summary, while traditional models and specialised solutions like Presidio have showcased strong capabilities, the custom Transformer Model stood out, reinforcing the transformative power and efficiency of advanced transformer architectures in the domain of data anonymisation. Their potential to extract intricate patterns and generalize well positions them as the front-runners for demanding tasks such as anonymisation.

\section{Conclusion \& Future Work}
\label{Conclusion}

This work ventures into investigating various machine learning models for the purpose of anonymization and provides a comparative study of results, offering significant insights. From our results, it is evident that the Transformer Model and CRF achieve superior performance in terms of precision, recall, and F1 score, with the Transformer Model slightly edging out in terms of recall. While models like GPT2, BERT, Presidio, and LSTM also showcased commendable performance, there remains room for optimisation. This is perceived as the difference for models adapted on highly specific tasks like anonymisation, like CRF and Transformer, in contrast with more generic models, GPT2, BERT, ELECTRA, that have been trained for generic tasks and fine-tuned for NER recognition. In future work, there is potential to explore ensemble techniques, combining the strengths of multiple models to enhance anonymisation performance further. 

\section*{Acknowledgment}
This project has received funding from the European Union’s Horizon Europe research and innovation programme under grant agreement No. 101070181 (TALON).

\bibliographystyle{IEEEtran} 
\bibliography{refs}

\begin{thebibliography}{10}
\providecommand{\url}[1]{#1}
\csname url@samestyle\endcsname
\providecommand{\newblock}{\relax}
\providecommand{\bibinfo}[2]{#2}
\providecommand{\BIBentrySTDinterwordspacing}{\spaceskip=0pt\relax}
\providecommand{\BIBentryALTinterwordstretchfactor}{4}
\providecommand{\BIBentryALTinterwordspacing}{\spaceskip=\fontdimen2\font plus
\BIBentryALTinterwordstretchfactor\fontdimen3\font minus \fontdimen4\font\relax}
\providecommand{\BIBforeignlanguage}[2]{{%
\expandafter\ifx\csname l@#1\endcsname\relax
\typeout{** WARNING: IEEEtran.bst: No hyphenation pattern has been}%
\typeout{** loaded for the language `#1'. Using the pattern for}%
\typeout{** the default language instead.}%
\else
\language=\csname l@#1\endcsname
\fi
#2}}
\providecommand{\BIBdecl}{\relax}
\BIBdecl

\bibitem{Zhiheng2015}
Z.~Huang, W.~Xu, and K.~Yu, ``Bidirectional lstm-crf models for sequence tagging,'' 2015.

\bibitem{Qu2020}
Y.~Qu, P.~Liu, W.~Song, L.~Liu, and M.~Cheng, ``A text generation and prediction system: Pre-training on new corpora using bert and gpt-2,'' in \emph{2020 IEEE 10th International Conference on Electronics Information and Emergency Communication (ICEIEC)}, 2020, pp. 323--326.

\bibitem{Vakili2022}
\BIBentryALTinterwordspacing
T.~Vakili, A.~Lamproudis, A.~Henriksson, and H.~Dalianis, ``Downstream task performance of {BERT} models pre-trained using automatically de-identified clinical data,'' in \emph{Proceedings of the Thirteenth Language Resources and Evaluation Conference}.\hskip 1em plus 0.5em minus 0.4em\relax Marseille, France: European Language Resources Association, Jun. 2022, pp. 4245--4252. [Online]. Available: \url{https://aclanthology.org/2022.lrec-1.451}
\BIBentrySTDinterwordspacing

\bibitem{Catelli2021}
R.~Catelli, F.~Gargiulo, E.~Damiano, M.~Esposito, and G.~De~Pietro, ``Clinical de-identification using sub-document analysis and electra,'' in \emph{2021 IEEE International Conference on Digital Health (ICDH)}, 2021, pp. 266--275.

\bibitem{Tjong2003}
\BIBentryALTinterwordspacing
E.~F. Tjong Kim~Sang and F.~De~Meulder, ``Introduction to the {C}o{NLL}-2003 shared task: Language-independent named entity recognition,'' in \emph{Proceedings of the Seventh Conference on Natural Language Learning at {HLT}-{NAACL} 2003}, 2003, pp. 142--147. [Online]. Available: \url{https://www.aclweb.org/anthology/W03-0419}
\BIBentrySTDinterwordspacing

\bibitem{Siniosoglou2021_a}
I.~Siniosoglou, P.~Sarigiannidis, Y.~Spyridis, A.~Khadka, G.~Efstathopoulos, and T.~Lagkas, ``Synthetic traffic signs dataset for traffic sign detection \& recognition in distributed smart systems,'' in \emph{2021 17th International Conference on Distributed Computing in Sensor Systems (DCOSS)}, 2021, pp. 302--308.

\bibitem{Kotei2023}
\BIBentryALTinterwordspacing
E.~Kotei and R.~Thirunavukarasu, ``A systematic review of transformer-based pre-trained language models through self-supervised learning,'' \emph{Information}, vol.~14, no.~3, 2023. [Online]. Available: \url{https://www.mdpi.com/2078-2489/14/3/187}
\BIBentrySTDinterwordspacing

\bibitem{Pilan2022}
\BIBentryALTinterwordspacing
I.~Pilán, P.~Lison, L.~Øvrelid, A.~Papadopoulou, D.~Sánchez, and M.~Batet, ``{The Text Anonymization Benchmark (TAB): A Dedicated Corpus and Evaluation Framework for Text Anonymization},'' \emph{Computational Linguistics}, vol.~48, no.~4, pp. 1053--1101, 12 2022. [Online]. Available: \url{https://doi.org/10.1162/coli\_a\_00458}
\BIBentrySTDinterwordspacing

\bibitem{Nikoletos2023}
S.~Nikoletos, S.~Vlachos, E.~Zaragkas, C.~Vassilakis, C.~Tryfonopoulos, and P.~Raftopoulou, ``Rog§: A pipeline for automated sensitive data identification and anonymisation,'' in \emph{2023 IEEE International Conference on Cyber Security and Resilience (CSR)}, 2023, pp. 484--489.

\bibitem{Pilan2023}
I.~Pilán, L.~Prévot, H.~Buschmeier, and P.~Lison, ``Conversational feedback in scripted versus spontaneous dialogues: A comparative analysis,'' 2023.

\bibitem{Patsakis2023}
C.~Patsakis and N.~Lykousas, ``Man vs the machine in the struggle for effective text anonymisation in the age of large language models,'' \emph{Scientific Reports}, vol.~13, 09 2023.

\bibitem{nasar2021named}
Z.~Nasar, S.~W. Jaffry, and M.~K. Malik, ``Named entity recognition and relation extraction: State-of-the-art,'' \emph{ACM Computing Surveys (CSUR)}, vol.~54, no.~1, pp. 1--39, 2021.

\bibitem{baigang2023review}
M.~Baigang and F.~Yi, ``A review: development of named entity recognition (ner) technology for aeronautical information intelligence,'' \emph{Artificial Intelligence Review}, vol.~56, no.~2, pp. 1515--1542, 2023.

\bibitem{raj2021anonymization}
A.~Raj and R.~D’Souza, ``Anonymization of sensitive data in unstructured documents using nlp,'' \emph{International Journal of Mechanical Engineering and Technology (IJMET)}, vol.~12, no.~4, pp. 25--35, 2021.

\bibitem{li2020survey}
J.~Li, A.~Sun, J.~Han, and C.~Li, ``A survey on deep learning for named entity recognition,'' \emph{IEEE Transactions on Knowledge and Data Engineering}, vol.~34, no.~1, pp. 50--70, 2020.

\bibitem{Taill2020}
B.~Taill{\'e}, V.~Guigue, and P.~Gallinari, ``Contextualized embeddings in named-entity recognition: An empirical study on generalization,'' in \emph{Advances in Information Retrieval}, J.~M. Jose, E.~Yilmaz, J.~Magalh{\~a}es, P.~Castells, N.~Ferro, M.~J. Silva, and F.~Martins, Eds.\hskip 1em plus 0.5em minus 0.4em\relax Cham: Springer International Publishing, 2020, pp. 383--391.

\bibitem{Kotevski2022}
\BIBentryALTinterwordspacing
D.~P. Kotevski, R.~I. Smee, M.~Field, Y.~N. Nemes, K.~Broadley, and C.~M. Vajdic, ``Evaluation of an automated presidio anonymisation model for unstructured radiation oncology electronic medical records in an australian setting,'' \emph{International Journal of Medical Informatics}, vol. 168, p. 104880, 2022. [Online]. Available: \url{https://www.sciencedirect.com/science/article/pii/S1386505622001940}
\BIBentrySTDinterwordspacing

\end{thebibliography}

\end{document}